# Robust Combination of Local Controllers

**Carlos Guestrin**
Department of Computer Science
Stanford University
Stanford, CA 94305-9010
guestrin@cs.stanford.edu

**Dirk Ormoneit**
Department of Computer Science
Stanford University
Stanford, CA 94305-9010
ormoneit@cs.stanford.edu

## Abstract

Finding solutions to high dimensional Markov Decision Processes (MDPs) is a difficult problem, especially in the presence of uncertainty or if the actions and time measurements are continuous. Frequently this difficulty can be alleviated by the availability of problem-specific knowledge. For example, it may be relatively easy to design controllers that are good locally, though having no global guarantees. We propose a nonparametric method to combine these local controllers to obtain globally good solutions. We apply this formulation to two types of problems: motion planning (stochastic shortest path problems) and discounted-cost MDPs. For motion planning, we argue that only considering the expected cost of a path may be overly simplistic in the presence of uncertainty. We propose an alternative: finding the minimum cost path, subject to the constraint that the robot must reach the goal with high probability. For this problem, we prove that a polynomial number of samples is sufficient to obtain a high probability path. For discounted MDPs, we consider various problem formulations that explicitly deal with *model uncertainty*. We provide empirical evidence of the usefulness of these approaches using the control of a robot arm.

## 1 Introduction

Planning is a central problem in artificial intelligence. In robot motion planning, for example, one is faced with the task of finding a collision-free path between an initial and a final configuration of a robot. For example, in a chemical plant, we may need to control the flow of chemicals to maintain certain characteristic reactions balanced. Markov Decision Processes (MDPs) have been used extensively as a framework for tackling such planning problems. Unfortunately, planning problems are frequently intractable due to the high dimensionality of the underlying system[1]. Many of the potential applications of planning algorithms are further complicated by the presence of uncertainty. The control of the robot may be imprecise or the sensors measuring chemical levels may be noisy. Furthermore, the state and the action spaces of MDPs are frequently continuous; few algorithms can deal with this situation.

On the other hand, problem specific characteristics can sometimes be exploited to tackle high dimensionality and continuous spaces. For instance, it is often possible to design simple controllers that work well locally, but have no global performance guarantees. In motion planning, one can use potential fields to design a good controller that can locally avoid obstacles well even in the presence of uncertainty, but cannot overcome local minima, preventing the robot from reaching a global goal [7]. Other examples of local controllers include PID-controllers, subsumption architectures and other user-designed local heuristics. A straightforward approach to simplify the solution of difficult planning problems is to combine several of these local controllers in a globally optimal fashion.

Of course, the general idea of combining local controllers is far from novel. For example, Hauskrecht et al. [4] as well as Parr [8] consider the optimal combination of local controllers in discrete MDPs. In this paper, we propose an approach that is suitable for *continuous* state spaces. We associate controllers with regions of the state space around anchor points, called *milestones*. The domain of the local controllers is the Voronoi partition implied by the milestones. We call this procedure *nonparametric* combination of local controllers, in analogy to nonparametric estimation. We describe algorithms to compute a global strategy from this type of policy representation in two types of problems: motion planning (stochastic shortest path problems) and discounted-cost MDPs. Our algorithm for motion planning is related to the probabilistic roadmap algorithm proposed for deterministic problems [6, 5]. Other algorithms for motion planning under uncertainty, e.g., *Preimage Backchaining* [7], have mainly focused on two dimensional problems and would be difficult to extend to higher dimensions. The MDP community has also studied this problem as the *stochastic shortest path* problem [1], however, most algorithms focus on discrete cases.

---
[1] Motion planning, for example, is known to be PSPACE-hard [9].

A particular focus is on the "robustness" of the solution: if the transition probabilities of the MDP must be estimated from data, there is *model uncertainty* in addition to the uncertainty inherent to the MDP. In the motion planning case, we suggest algorithms that account for model uncertainty by minimizing the path cost subject to a constraint on the minimum allowable probability of success. Experimental results illustrate the qualitative and quantitative characteristics of this type of optimization. Also, for the motion planning case, it is possible to prove that in expansive spaces, a concept used previously in motion planning [5], it is possible to obtain a high probability path with a polynomial number of samples. For discounted-cost MDPs, we suggest that it is also possible to nonparametrically combine local controllers. We propose a formal framework to account for model uncertainty in this case, and review algorithms to deal with this type of uncertainty.

## 2 Motion planning using local controllers

In this section, we present an application of the concept of nonparametric combinations of local controllers to motion planning. There are several sources of uncertainty in this case, e.g., the control of the robot may not be precise, the position of the obstacles may not be known exactly, there may be moving obstacles for which the motion is uncertain, etc. In the presence of uncertainty, it is hence crucial to find a path from a starting to a goal configuration that has a high probability of success; that is, a high probability of reaching the goal without collisions. We present an algorithm to compute the path that is effective in this sense in Section 2.1. A conflicting goal is to minimize the costs of the path, e.g. the distance traveled, time and energy spent on the way from the initial to the goal configuration. A conservative mindset suggests placing more emphasis on the probability of error than on the cost of a path. Hence a sensible approach is to minimize the costs using the probability of error as a robustness constraint (Section 2.2). Section 2.3 gives a theoretical result regarding the probability of finding a path with low probability of error. Prior to discussing these results, we first formalize the basic building blocks for a motion planning problem under uncertainty with local controllers:

**Motion planning problem:** A *deterministic* motion planning problem is defined by a compact *configuration space* $\mathcal{C}$, that is the set of all possible configurations $x$ of the robot, and by an open subset $\mathcal{F} \subseteq \mathcal{C}$ that is *free* in the sense that the configurations in $\mathcal{F}$ are physically possible, thus excluding collisions with obstacles, self collisions, etc. As we mentioned above, the objective is to find a collision free path between a starting configuration $x^S \in \mathcal{C}$ and a goal configuration $x^G \in \mathcal{C}$ which minimizes the cost $c$ of the path.

We modify this definition of the deterministic motion planning problem to account for uncertainty as follows: It may not be possible to obtain a path that is guaranteed to be collision free in the presence of uncertainty. Instead we content ourselves with a path that has a high probability of success. This is where the notion of local controllers becomes important. We can design local controllers that will take the robot from one configuration (milestone) to another nearby configuration (milestone) with high probability. Thus, we redefine the objective as one of finding a set of milestones $\mathbf{X} \equiv (x_0, \ldots, x_n)'$, $x_i \in \mathcal{C}$, such that $x_0 = x^S$, $x_n = x^G$ and the local controller can take the robot from $x_i$ to $x_{i+1}$ with high probability. Below we present an algorithm to address this problem, after defining a local controller.

**Local controllers:** A local controller $a$ has as parameters a starting configuration $x_i$ and a goal configuration $x_j$. The objective is to steer from $x_i$ towards the local goal $x_j$. Here the controllers may be either smart and try to avoid obstacles (or even model uncertainty locally) or they may be simple and, say, try to maintain a straight line path to the goal.

The local controller evaluates whether it can (locally) connect two configurations $x_i$ and $x_j$ using a *generative model*. That is, it repeatedly starts at $x_i$ and simulates the local control as it tries to reach $x_j$. Each simulated transition $t$ has associated with it a cost $c_{i,j}^t$, e.g., the distance traveled by the robot along the path from $x_i$ to $x_j$. There are two termination conditions for each simulation: First, the controller may hit an obstacle or the search may exceed a time limit which corresponds to a failure of the local connection. Second, it may reach an "endgame region" around the goal $x_j$ in which case the controller stops successfully and returns the accumulated cost $c_{i,j}^t$. In unsuccessful trials $c_{i,j}^t = 0$.

**Error probabilities:** We carry out a fixed number $T$ of these trials and let $T_{success}$ be the number of successful trials. Let $\hat{c}_{i,j} \equiv 1/T_{success} \sum_{t=1}^{T} c_{i,j}^t$ and $\hat{p}_{i,j} = T_{success}/T$ denote the average costs and the empirical frequency of success of this experiment, respectively. As we mentioned above, $\hat{p}_{i,j}$ may deviate from the true transition probability $p_{i,j}$ due to the random sampling, and it is important to account for this model uncertainty when searching robust paths. To express the model uncertainty formally, note that $\hat{p}_{i,j}$ can be viewed as a random variable, resulting from averaging independent Bernoulli variables (simulation outcomes) with probability of success equal to the (unknown) true $p_{i,j}$. Hence $\hat{p}_{i,j}$ is — except for the normalization — binomially distributed with unknown parameters. We can approximate this binomial distribution with the normal distribution $N(p_{i,j}, p_{i,j} - p_{i,j}^2)$ asymptotically. The straightforward way to define a lower bounds on $\hat{p}_{i,j}$ with reliability of at least say $\gamma$ is to consider the corresponding tail probability of the normal distribution, $\Phi^{-1}(\gamma)$, rescaled by the root of the empirical variance $\hat{\sigma}_{i,j}^2 = \frac{1}{T-1}\left[T_{success}(1 - 2\hat{p}_{i,j}) + T\hat{p}_{i,j}^2\right]$. The resulting (lower) bound for $p_{i,j}$ is:

$$\underline{p}_{i,j} = \max\{\hat{p}_{i,j} - \hat{\sigma}_{i,j}\Phi^{-1}(\gamma), 0\}.$$

We consider the lower bound because we are interested in guaranteeing good performance in the worst-case scenario. Naturally, this perspective may not lead to any feasible solution in the case where we have only very few trials for each connection.

### 2.1 Finding the path with the least probability of error

We can now generate a graph $\langle V, E \rangle$, or roadmap, that succinctly represents how the local controller can act globally. The vertices of this graph are the milestones. There is an edge between two milestones $x_i$ and $x_j$ if $p_{i,j} \neq 0$ and each edge is augmented with the values $p_{i,j}$ and $c_{i,j}$. The algorithm proceeds as follows:

1. Sample $n - 1$ milestones $x_1, \ldots, x_{n-1}$ uniformly from $\mathcal{F}$ by taking samples uniformly from $\mathcal{C}$ and rejecting those samples that result in collisions with certainty.
2. Try to connect each milestone $x_i$ to its $k$ nearest neighbors using the local controller as described above. This gives the cost $c_{i,j}$ and the probability of failure $p_{i,j}$.
3. For each new query for a path from a starting position $x^S$ to a goal position $x^G$, include $x^S$ and $x^G$ into the set of milestones using the indices 0 and $n$, repsectively. That is, $\mathbf{X} \equiv (x_0, \ldots, x_n)'$ with $x_0 \equiv x^S$ and $x_n \equiv x^G$ as described above. Also, use the local controller to detmine $p_{0,j}$ and $p_{i,n}$.
4. Determine the shortest path between $x_0$ and $x_n$ in the graph induced by $\mathbf{X}$ using $-\log(p_{i,j})$ as the edge weights.
5. Apply local controller along the shortest path.

Thus, the algorithm we present finds the set of milestones that gives the highest probability path in the graph from start to goal. As we will prove in Sec. 2.3, it is possible to obtain (with high probability) a path that has high success probability, even with a polynomial number of milestones. Another issue is the cost of the resulting path. In the next section we present an approptiate tradeoff between the costs and the probability of success.

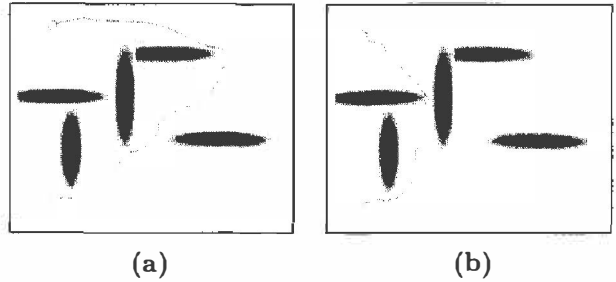

Figure 1: Single robot example: (a) setting success probability $p_{min} = 0.99$ yields a path of length 1.75; (b) lowering $p_{min}$ to 0.51 the path is 1.08 long.

### 2.2 The shortest path subject to robustness constraints

In the previous section, we considered the problem of finding a viable connection between two locations in the state space with a low probability of failure. In most practical problems an additional requirement is that the path followed should be "cheap" in a suitable sense, e.g., it is typically desirable to move a robot arm to its goal position without unnecessary detours.

More formally, we need to take into account the cost $c_{i,j}$ associated with the motion between locations $i$ and $j$ in addition to the probability of failure $p_{i,j}$ introduced above. This extension immediately leads to a difficulty with our problem definition: clearly, the tasks of finding a path with a low probability of error and that of finding a path with low costs may be contradictory. The shortest path must usually run very close to obstacles, e.g., when turning a corner the robot must almost touch the corner to make the shortest path. However, paths closer to obstacles usually imply higher probability of colliding with them. This is illustrated in Fig. 1, where we are interested in planning a path for two degrees of freedom (dof) holonomic circular robot. However, the positions of the obstacles are not perfectly known, e.g., they could have been measured by a noisy sensor. The noise in this example is gaussian and the shaded areas are a graphical representation of this distribution. In Fig. 1(a), we try to find a path with very high probability of success. This is a conservative path that chooses long but safe route. On the other hand, in Fig. 1(b), lowering the constraint on the success probability, we obtain a much shorter path taking a "risky" passage between obstacles.

There are several ways to tradeoff costs and probability of success mathematically. From a conservative viewpoint, it is more important to reach the goal at all than to reach it cheaply. Hence we cast our algorithm as a constrained optimization problem, where the objective is to minimize the path cost and the constraint is that the path chosen must lead to an overall probability of success of at least some pre-specified value $p_{min}$. This problem formulation is natural in many practical cases. For example, in motion planning it is very important that the robot arm does not crash into any obstacle and reaches the goal whereas it is less relevant whether or not the path chosen is slightly suboptimal in terms of its length.

Note that the optimization we propose is different from the usual objective in MDP algorithms, which is to minimize the expected cost. In the usual MDP formulation, one would have to define, in addition to the cost function associated with the problem, a cost for hitting obstacles and another (reward) for reaching the goal. The tradeoff becomes implicit and difficult to control, forcing the user to tune the cost function, balancing the relative weights of these three quantities.

Our new constrained optimization problem can be cast as a restricted shortest-path problem. We modify the fourth step of the algorithm of Sec. 2.1 to incorporate this new optimality criteria. Theoretically it is well-known that restricted shortest-path problems are NP-hard, but fully polynomial approximation schemes (FPAS) have been developed for their approximate solution [3]. In this section, we describe an algorithm that is *not* FPAS but that is easy to implement and worked very well in our experiments.

We discretize the range $[p_{min}, 1]$ into $S+1$ values using $q(s) := (p_{min})^{s/S}$ for $s = 0, \ldots, S$. Intuitively we think of the value $V(s, i)$ as the minimum cost to reach $x_i$ from $x_0$ with a probability of success of at least $q(s)$. For simplicity of exposition, we will assume that all success probabilities are strictly smaller than one ($p_{i,j} < 1$).[2] The algorithm computes the value function at each vertex of the graph by a simple dynamic programming algorithm:

1. $V(s, 0) := 0$ for $s = 0, \ldots, S$.
2. $V(0, i) := \infty$ for $i = 1, \ldots, n$.
3. For $s := 1, \ldots, S$; For $j := 1, \ldots, n$:

$V(s, j) :=$
$\min \left\{ \begin{array}{l} V(s-1, j) \\ \min_{k \in parents(j)} \{V(\lfloor s - s_{k,j} \rfloor, k) + c_{k,j}\} \end{array} \right\}$,
where $s_{k,j} := S \log p_{k,j} / \log p_{min}$.

In practice, we suggest applying first the algorithm described in the previous section to find the maximal probability of success. If this is at an acceptable level, then we can set $p_{min}$ to a value less than or equal to this maximal probability and run the constrained optimization algorithm. If the probability of success is not at an acceptable level, we add more milestones to the graph or use a better local controller. The remaining question is how many milestones are needed to obtain high success probability paths. In the next section, we provide a formal answer to this question, which implies that a high probability path can be obtained with a polynomial number of samples.

### 2.3 Theoretical analysis

In this section, we analyze the algorithm presented in Sec. 2.1 and show that it is possible to obtain a path with high probability of success using a polynomial number of samples. There are two key issues to be considered: first, how many milestones are needed to make it sufficiently likely that the graph contains a path between $x^S$ and $x^G$; second, what is the success probability of the resulting path. As we will show, these two issues can be addressed theoretically by extending the analysis of Hsu, Latombe, and Motwani [5] for the deterministic motion planning case.

The analysis of these authors shows that if the configuration space is "expansive", it is possible to find a path from the starting point to the goal using a polynomial number of samples [5, Hsu et al.]. In the rest of this section, we extend their formal framework to problems with uncertainty.

First, we extend some definitions. Let a $\dot{p}$-good expansive space $(\varepsilon, \alpha, \beta, \dot{p})$ under some local controller $a$ be defined as follows: By analogy to the deterministic case, we let $\mathcal{R}_{\dot{p}}(x)$ denote the region reachable by the local controller starting at $x$, i.e., the set of points that can be reached from the configuration $x$ using the local controller $a$ with probability of success of at least $\dot{p}$. Let $\mu(A)$ denote the volume of the set $A$. Now, we need to extend the concept of $\varepsilon$-goodness, which implies that the controller can reach at least some proportion of the free space with high probability:

**Definition 2.1** *Let $\varepsilon \in (0, 1]$ be a constant. A free space $\mathcal{F}$ is $\varepsilon$-$\dot{p}$-good if for every connected component $\mathcal{F}' \subseteq \mathcal{F}$, for every $x \in \mathcal{F}'$, $\mu(\mathcal{R}_{\dot{p}}(x)) \geq \varepsilon \cdot \mu(\mathcal{F}')$.*

Next, we need to define the *lookout* of a set $S$:
$LOOKOUT^{\dot{p}}_{\beta}(S) = \{x \in S | \mu(\mathcal{R}_{\dot{p}}(x) \setminus S) \geq \beta \cdot \mu(\mathcal{F}' \setminus S)\}$.
Intuitively, this quantity measures the proportion of $S$ that can reach many points outside $S$. Finally, we can extend the concept of $(\alpha, \beta, \dot{p})$-expansiveness:

**Definition 2.2** *Let $\alpha \in (0, 1]$ and $\beta \in (0, 1]$ be constants, a free space $\mathcal{F}$ is $(\alpha, \beta, \dot{p})$-expansive if for every connected component $\mathcal{F}' \subseteq \mathcal{F}$, for every $S \subseteq \mathcal{F}'$, $\mu(LOOKOUT^{\dot{p}}_{\beta}(S)) \geq \alpha \cdot \mu(S)$.*

---
[2]Probability one transitions can be dealt with by adding an extra loop to the algorithm.

These extensions of the original definitions of expansive spaces allow us to prove the following theorem:

**Theorem 2.3** *For any $\gamma \in (0, 1]$, let a roadmap be generated from a set $M$ of $2\lceil 8\ln(8/\varepsilon\alpha\gamma)/\varepsilon\alpha + 3/\beta\rceil + 2$ independent uniformly sampled milestones from the free space $\mathcal{F}$; then, with probability at least $1 - \gamma$, the roadmap will contain a path between any two milestones in the same connected component with probability of success of at least $\tilde{p}^{3/\beta+1}$.*

**Proof sketch:** the arguments are similar to those for the deterministic motion planning case of Hsu et al. [5]. Our extension of the definition of expansiveness to the stochastic case allows us, using similar arguments, to obtain results equivalent to Lemmas 1 and 2 in [5]. Our final result, which is analogous to the deterministic case addressed by Hsu et al. in Theorem 1, uses a similar *linking sequences* argument [5]. As a result of this argument, any two milestones $q_0$ and $q_{m+1}$ in the same connected component will, with probability $1 - \gamma$, be connected by a path that contains at most $m \leq 3/\beta$ milestones $\{q_0, q_1, \ldots, q_m, q_{m+1}\}$. Furthermore, for each $i$, $q_{i+1} \in \mathcal{R}_{\tilde{p}}(q_i)$, i.e., the local controller must be able to navigate from $q_i$ to $q_{i+1}$ with success probability of at least $\tilde{p}$. Implying that the complete path has success probability of at least $\tilde{p}^{3/\beta+1}$. ∎

Note that we do not impose any explicit smoothness constraints on the transition density in this work. This is by contrast to an algorithm by Rust [10] for the approximate solution of continuous state, discrete actions and discrete time discounted MDPs, which depends crucially on the Lipschitz continuity of the transition density. Rust's algorithm is more effective when the Lipschitz constant is small, i.e., there is a lot of randomness in the system. On the other hand, our algorithm becomes more efficient as the controllers become more effective (deterministic) locally. We believe that, in most practical robotics applications, local controllers are close to deterministic, thus, making our algorithm more suitable.

In summary, in the context of expansive spaces, it is possible to obtain, with high probability, a path between two milestones using a number of samples that is polynomial in $(1/\varepsilon, 1/\alpha, 1/\beta, \ln 1/\gamma)$. In particular, the number of samples needed depends only on the number of "critical" dimensions of the problem, as was argued by Hsu et al. [5]. This number is frequently relatively small in practice, allowing for a good performance of the algorithm even in relatively high-dimensional spaces.

## 3 Discounted MDPs

In the previous section, we suggested a robust algorithm for motion planning using local controllers that minimizes the cost of a path subject to a reliability constraint. The idea of using local controllers can be very useful not only for motion planning, but also for discounted MDPs.

As an example, consider the task of riding a bicycle. Here we may have different controllers at our disposal that show different degrees of reliability in different regions of the state-space. For instance, one controller may be better suited for "normal" riding scenarios and another controller may work well in "emergency situations". A globally optimal strategy would identify these scenarios and assign the local controllers accordingly to achieve an overall good performance.

### 3.1 Evaluating combinations of local controllers

For discounted MDPs, we assume we have several different local controllers, indicated by the variable $a$. The task is then to choose one controller at each state in a way that minimizes the overall total expected discounted cost. In this problem formulation, future rewards are discounted using a discount factor $\alpha \in [0, 1)$.

The use of local controllers (macro-actions) has been explored in discrete MDPs, e.g. [4, 8]. The algorithm by Hauskrecht et al. assumes some partitioning of the state space into regions, it is then able to generate local policies within regions and combines them into a global policy. Unfortunately, extending this concept to continuous problems is difficult. The boundaries between regions can be complicated to describe. Furthermore, in a $d-dimensional$ continuous state space, the boundaries define a $(d-1) - dimensional$ manifold which is difficult to deal with computationally. Our approach circumvents this problem by implicitly defining the boundaries using a set of milestones: We assume that a set of local policies is given, and we divide the state space into local regions using a set of milestones $\{x_0, \ldots, x_n\}$, as in Section 2. These regions are defined in terms of a Voronoi partition of the state space; that is, each location is assigned to the nearest milestone. Forthermore, each region can employ one of several local controllers and the objective is to find the optimal assignment of controllers to regions.

We explore the connectivity between the milestones by simulating each controller $a$ starting from uniformly generated starting points in $\mathcal{F}$. Each point is associated with its nearest milestone, $x_i$, and the simulation is terminated as soon as the trajectory is closer to another milestone $x_j$ than to $x_i$. We let $I(i,j)$ be the indicator that a simulation starting in the vicinity of $x_i$ terminates in the vicinity of $x_j$, and we let $\tau$ denote the corresponding "stopping time". Our goal is to approximate the discounted transition probability $P_a^\alpha(i,j) = E_\tau[\alpha^{\tau-1} I(i,j)]$, where the term $\alpha^{\tau-1}$ accounts for the different stopping times. For this pur-pose, suppose we have $T_i$ trial simulations that start in the vicinity of each state $x_i$ for each controller $a$. Let $\tau_{i,j}^k$ be the stopping time of the $k$th simulation starting from $x_i$ that ended at $x_j$. Furthermore, let the discounted cost associated with the path generated during the $k$th trial be denoted as $c_i^k$. Then straightforward estimates of the transition costs and of the discounted transition probabilities are defined according to:

$$\hat{c}_a(i) = 1/T_i \sum_{k=1}^{T_i} c_i^k; \text{ and } \hat{P}_a(i,j) = 1/T_i \sum_{\tau^k} \alpha^{\tau_{i,j}^k - 1}.$$

Note that these estimates are heuristic, because we stop the simulation at the boundary of the neighboring cell, as described above. As a consequence, there arises a bias because we ignore information regarding the motion *within* the neighboring cell.[3]

In our second step, we use the estimates $\hat{P}_a$ and $\hat{c}_a$ to determine the optimal assignment of controllers to regions. A straightforward approach to compute the solution of the discrete-state, discrete-action MDP defined by the estimates $\hat{P}_a$ and $\hat{c}_a$ for this purpose. However, like in the motion planning case it is very important to account for the random variation in $\hat{P}_a(i,j)$ due to the random simulation. In the next section, we suggest approaches to deal with this type of *model uncertainty* in the context of discounted MDPs.

### 3.2 Robust solution of the approximate MDP

We emphasized above that a drawback of any simulation based algorithm for planning under uncertainty, e.g. the algorithms of Sections 2 and 3.1, is the *model uncertainty* that may produce misleading results. It is important to account for this uncertainty as it was shown in Sections 2.1 and 2.2 for motion planning. The case of discounted MDPs is mathematically more involved, and we give an overview of useful algorithms in this section.

Algorithms for discounted MDPs with model uncertainty have been discussed in some detail in the operations research and computer science literatures. A comprehensive mathematical treatment of MDPs with "imprecise transition probabilities" can be found in a paper by White and Eldeib [11]. These authors are concerned with the case where we have a number of linear constraints on the $i$th transition probability vector $\hat{P}_a(i,\cdot)$. Givan et al. [2] specialize this method to express model uncertainty in the form of elementwise bounds on the parameters $\hat{c}_a(i)$ and $\hat{P}_a(i,j)$. They also derive modifications of the policy iteration and value iteration algorithms for this generalized class of MDPs and highlight their relationship to stochastic games.

Applied to the current context, their algorithm amounts to defining elementwise bounds $\underline{P}_a(i,j)$ and $\overline{P}_a(i,j)$ by analogy to Section 2, as well as corresponding bounds on the cost function, $\underline{c}_a(i)$ and $\bar{c}_a(i)$. Formally, these bounds translate into the sets of feasible transition matrices and cost vectors $\mathcal{P}_a = \{P_a | \underline{P}_a \leq P_a \leq \overline{P}_a\}$ and $\mathcal{C}_a = \{c_a | \underline{c}_a \leq c_a \leq \bar{c}_a\}$, where the '$\leq$' is to be interpreted elementwise. In addition, they define *interval value functions* in terms of elementwise upper and lower bounds $\overline{V}(i)$ and $\underline{V}(i)$, respectively. These value functions are updated recursively in a variant of the value iteration algorithm as follows:

$$\overline{V}'(i) := \min_a \max_{P_a \in \mathcal{P}_a} \left\{ \bar{c}_a(i) + \sum_j P_a(i,j)\overline{V}(j) \right\};$$

$$\underline{V}'(i) := \min_a \min_{P_a \in \mathcal{P}_a} \left\{ \underline{c}_a(i) + \sum_j P_a(i,j)\underline{V}(j) \right\}.$$

This algorithm converges to a unique fixed point that characterizes the optimal solution, defined using a suitable notion of robustness [2].

In practice, a potential difficulty with the algorithm by Givan et al. is that, depending on the bounds on $P_a$ and $c_a$, the range of possible value functions that are consistent with $\overline{V}(i)$ and $\underline{V}(i)$ may be too big and, as a consequence, these bounds may not be sufficiently informative to discriminate between good and bad policies. Hence it is important to define the sets $\mathcal{P}_a$ and $\mathcal{C}_a$ in a way that is as restrictive as possible in practice.

In particular, the elementwise bounds introduced above may be too loose in many applications, and there are many ways to arrive at tighter constraints at the cost of additional mathematical sophistication and computational effort. From a statistical perspective, a natural way to define those constraints is by formulating likelihood functions that involve $c_a$ and $P_a$ as parameters, and then to use the sublevel sets of the likelihood. While the constraints derived in this manner are frequently convex — and hence amenable to numerical optimization — they can still be difficult to deal with computationally. A reasonable compromise is to derive spherical bounds using a *Laplace approximation* of the likelihood: We consider the sublevel sets of a second-order Taylor expansion of the log-likelihood, which can be described as spheres with respect to the inverse Fisher information matrix $\Omega$. In the case of the transition matrix $P_a$, the resulting constraints on the $i$th row can be written in the form:

$$P_a(i,\cdot) \in \mathcal{P}_{a,i} = \left\{ \hat{P}_a(i,\cdot) + \delta w \mid \|w\|_\Omega \leq 1, w'\mathbf{1} = 0 \right\},$$

---
[3]Hauskrecht et al. account for this difficulty by solving a smaller, but still $d$-dimensional, subproblem within each cell. While theoretically appropriate, this approach may be extremely demanding in practice if $d$ is large.

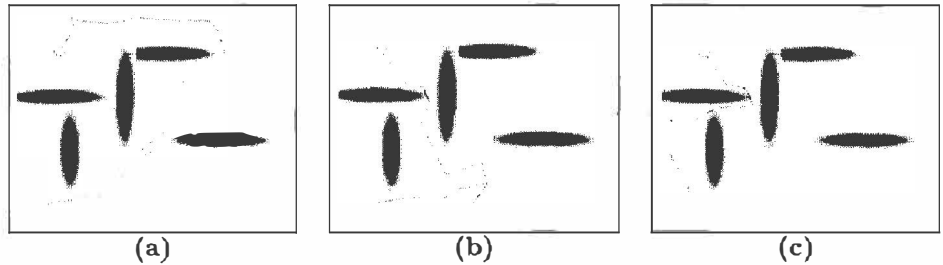

Figure 2: Two robots, roadmap with 2000 milestones, considering 10 nearest neighbors: (a) most conservative path ($p_{min} = 0.99$); (b) as more risk is allowed ($p_{min} = 0.54$) one robot enters uncertain narrow passage; (c) both robots use narrow passage in riskiest situation ($p_{min} = 0.13$).

where $||z||_\Omega = \sqrt{z'\Omega z}$ is the $\Omega$-weighted Euclidian norm.

A robust algorithm is to find a policy that is optimal with respect to the most adverse transition probability in this class, by analogy to the interval value iteration algorithm described above. For this purpose, note that the equation above amounts to a set of linear equality and quadratic inequality constraints. Evaluating $\max_{P_a \in \mathcal{P}_a} \left\{ \overline{c}_a(i) + \sum_j P_a(i,j) \overline{V}(j) \right\}$ in the update equation for $\overline{V}'(i)$ in the interval value iteration algorithm thus requires solving a linear program with (convex) quadratic constraints. The iterated solution of this program gives a robust estimate of the value function. It can be used to derive an approximately optimal strategy in a straightforward manner.

To summarize, our overall algorithm consists of first sampling a set of milestones, then estimating the transition probabilities between each neighboring region using simulation, and finally solving a "robust" MDP to derive a suitable global strategy. There are several ways to define the robust MDP that vary in their algorithmic complexity and the expressiveness of their solution. A useful heuristic is to start from relatively simple (elementwise) bounds in practice, and to resort to more sophisticated methods only as additional precision is needed.

## 4 Experiments

In this section, we present an implementation of the motion planning algorithm described in Section 2. The test cases involve uncertainty in the position of obstacles. This type of uncertainty could result from noisy sensor data, for example. The noise in these examples is gaussian and the shaded areas are a graphical representation of the uncertainty. Since it is difficult to represent paths in a paper, we strongly encourage the reader to see animated paths and other examples on the project website: robotics.stanford.edu/~guestrin/Research/RobustLocalControl/.
We have not tried to do any local optimization or smoothing of the paths in these experiments. This would make paths somewhat shorter and more natural, but might hide certain characteristics of the algorithm. To navigate between nearby milestones, $x_i$ to $x_j$, we used a simple local controller that attempts to traverse a straight line path from $x_i$ to $x_j$.

The first example, shown in Fig. 2, deals with the centralized control of two circular holonomic robots. The goal is to minimize the total costs while avoiding the obstacles at the same time. Accordingly, the paths in Fig. 2 have qualitatively different behaviors as we relax the constraints on the minimum allowable success probability. For the highest success probability (Fig. 2(a)), both robots must take a long path around the obstacles to avoid regions of likely collisions. As we relax the constraint on the success probability (Fig. 2(b)), the planner is able to route one robot through a risky (high uncertainty in the position of obstacles) short cut, making the overall path length shorter. Finally, as we relax the constraints further, both robots take the risky short cut. The quantitative differences are:

| Example | Path length | Prob. success |
|---|---|---|
| Fig. 2(a) | 3.53 | 0.99 |
| Fig. 2(b) | 2.79 | 0.54 |
| Fig. 2(c) | 1.53 | 0.13 |

The second example concerns the control of a 5 degrees of freedom (dof) robot arm. The most probable path (Fig. 3) still has to pass through an area of high uncertainty; this is a narrow passage that cannot be avoided. However, it is able to avoid other uncertain areas. As we relax the constraint on the failure probabilities, we can obtain shorter paths (Fig. 4) at the cost of entering areas of high risk. Quantitatively:

| Example | Path length | Prob. success |
|---|---|---|
| Fig. 3 | 10.07 | 0.95 |
| Fig. 4 (a) | 9.23 | 0.81 |
| Fig. 4 (b) | 7.81 | 0.60 |

To illustrate the degree of uncertainty present in the problem, the histogram in Fig. 5 shows the distribution in values of $p_{i,j}$ determined during the simulation. Note that 70% of the edges have probability of success less than 1, demonstrating the relevance of dealing with uncertainty explicitly.

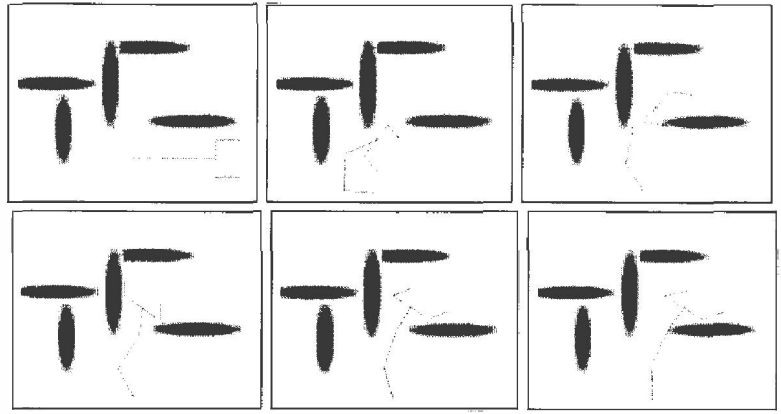

Figure 3: Five dof arm, roadmap with 1000 milestones, considering 10 nearest neighbors: highest probability path ($p_{min} = 0.95$) requires traversal risky passage ($3^{rd}$ image) and detour ($4^{rd}$ image).

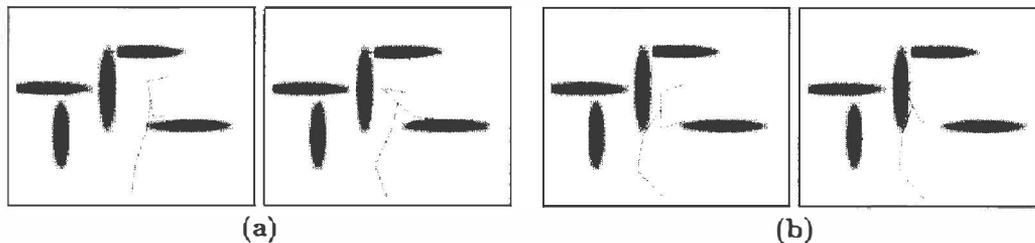

Figure 4: (a) shorter path ($p_{min} = 0.81$) requires entering risky area two more times; (b) much shorter path is obtained by entering more areas of high collision probabilities ($p_{min} = 0.60$).

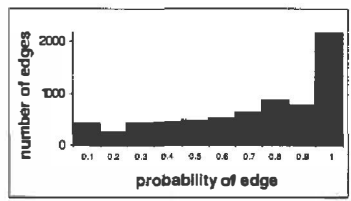

Figure 5: Distribution of edge probabilities $p_{i,j}$.

## 5 Discussion

We presented new algorithms for planning under uncertainty in continuous state and action spaces, which are based on the combination local controllers. For motion planning, we argue that expected cost — the usual objective function for planning using MDPs — may be inappropriate in situations where the cost of a path is dominated by the need to eliminate the probability of error. We propose an alternative, minimizing path cost subject to a constraint on the minimum acceptable probability of reaching the goal. This concepts leads to an algorithm that is guaranteed to yield a robust (high success probability) solution path. Experiments show our problem formulation successfully trades off risk and reward in two planning scenarios.

Also for discounted MDPs model uncertainty is an important issue, and we suggest a similar approach to combine local controllers robustly in this context. In detail, we suggest a way to quantify the model uncertainty mathematically and review techniques to solve the resulting robust MDP, varying in mathematical sophistication and expressiveness of the solution.

## Acknowledgments

We would like to thank Daphne Koller for help and support. This work was supported by the ONR under the MURI program "Decision Making Under Uncertainty", by the ARO under the MURI program "Integrated Approach to Intelligent Systems", and by the Sloan Foundation.